\documentclass[twocolumn]{article}
\usepackage[margin=2cm]{geometry}
\usepackage{graphicx}
\usepackage[colorlinks=true, allcolors=blue, breaklinks]{hyperref}
\usepackage{float}
\usepackage{xurl}
\usepackage{blindtext}
\usepackage{setspace}
\usepackage{subfig}
\usepackage{todonotes}

\title{Auto deep learning for bioacoustic signals}
\author{Giulio Tosato\textsuperscript{1*}
\and Abdelrahman Shehata\textsuperscript{1}
\and Joshua Janssen\textsuperscript{1}
\and Kees Kamp\textsuperscript{1}
\and Pramatya Jati\textsuperscript{1}
\and Dan Stowell\textsuperscript{1,2}
}
\date{}

\usepackage{todonotes}

\begin{document}
\twocolumn[
  \begin{@twocolumnfalse}
    \maketitle
    \begin{center}
      \textsuperscript{1} Department of Cognitive Science and Artificial Intelligence, Tilburg University, Tilburg

      \textsuperscript{2} Naturalis Biodiversity Center, Leiden
      
      \textsuperscript{*} Corresponding author: g.tosato@tilburguniversity.edu
    \end{center}
    \bigskip
  \end{@twocolumnfalse}
]

\section*{Abstract}
This study investigates the potential of automated deep learning to enhance the accuracy and efficiency of multi-class classification of bird vocalizations, compared against traditional manually-designed deep learning models. Using the Western Mediterranean Wetland Birds dataset, we investigated the use of AutoKeras, an automated machine learning framework, to automate neural architecture search and hyperparameter tuning.
Comparative analysis validates our hypothesis that the AutoKeras-derived model consistently outperforms traditional models like MobileNet, ResNet50 and VGG16.  Our approach and findings underscore the transformative potential of automated deep learning for advancing bioacoustics research and models. In fact, the automated techniques eliminate the need for manual feature engineering and model design while improving performance. 
This study illuminates best practices in sampling, evaluation and reporting to enhance reproducibility in this nascent field. All the code used is available at \url{https://github.com/giuliotosato/AutoKeras-bioacustic}

\paragraph{}

\textbf{Keywords}: AutoKeras; automated deep learning; audio classification; Wetlands Bird dataset; comparative analysis; bioacoustics; validation dataset; multi-class classification; spectrograms.

\section{Introduction}

The composition of bird communities and fluctuations in specific species' populations serve as reliable indicators of an ecosystem's overall well-being (Tuggener et al., 2019). Through bioacoustic monitoring —a practice that records and analyzes natural sounds, including bird calls— ornithologists gain invaluable insights into avian populations, behaviors, and biodiversity. This acoustic information not only aids in identifying endangered or declining species but also offers critical data for conservation initiatives.
Bioacoustics research has historically relied on manual identification of bird species by experts but with the emergence of machine learning and large audio datasets, 
automatic recognition of sounds through deep learning has become feasible and popular (Stowell 2022).
However, building highly accurate deep learning models presents several challenges. Large labeled datasets with many examples per class are required, but collecting and annotating sufficient bird vocalization data is difficult and labor-intensive. The models themselves are complex, with many architectural design choices (e.g. neural network topology, hyperparameters) that must be optimized for a particular problem. This neural architecture engineering relies heavily on human expertise through trial-and-error testing.  Moreover, the computational demands for training these deep networks are substantial, usually necessitating the use of high-performance GPUs.
Together, these factors create a high barrier to entry for domain specialists like ecologists or conservationists looking to apply deep learning for bioacoustics. 

Recent work has investigated \textit{automated machine learning} or \textit{AutoML}, in which a meta-algorithm automatically optimizes over these choices. This provides a solution to automate model development and potentially outperform manual approaches while enabling domain experts with limited machine learning expertise to create high-quality models (Truong et al., 2019). 

Brown et al. (2021) formulate the problem of automating bioacoustics analysis as one of workflow construction. They aim to build full workflows by searching over different tasks like preprocessing, feature extraction, and classification. Their goal is to find combinations of tasks that maximize accuracy and efficiency for species detection in synthetic soundscapes. They treat workflow construction as a combinatorial optimization problem and use a particle swarm optimization algorithm guided by a neural network surrogate model to search the space of possible workflows. The surrogate model predicts the accuracy of workflows to guide the search. They test this approach on different soundscapes and compare it to algorithms like genetic algorithms. The focus is on constructing workflows from reusable tasks.

In contrast, our work focuses specifically on multi-class bird vocalization classification problem and aim to find the optimal neural network architecture and hyperparameters using neural architecture search with AutoKeras. AutoKeras automatically searches over deep learning models, tuning the architecture and hyperparameters
using network morphism (Wei, 2016).  Network morphism in AutoKeras is built upon Bayesian optimization to search for optimal neural network architecture and hyperparameters for efficient neural architecture search (Jin et al., 2019).The framework incorporates a neural network kernel and a tree-structured acquisition function optimization algorithm to explore the search space of possible deep models more effectively, which results in more accurate and efficient feature extraction.

Our central hypothesis is that an AutoKeras-derived model will demonstrate superior performance over manually-designed models like MobileNet, ResNet50 and VGG16 on a multi-class bird vocalization classification task.
To test this hypothesis, we use the publicly available Western Mediterranean Wetland Birds dataset containing vocalizations from 20 endemic species. Our methodology incorporates stratified sampling and data normalization for multi-class learning. Additionally, we underscore the critical need for a separate test set to properly evaluate model generalization, given AutoML’s reliance on the validation set for model selection.

Overall, our automated techniques eliminate manual feature engineering while improving performance, illuminating the  potential to advance bioacoustics research and models. The study provides key insights into sampling, evaluation and reporting to enhance reproducibility in this nascent field.

\section{Methods}
\subsection{Dataset}
The Western Mediterranean Wetland Birds dataset was used in this study, consisting of 201.6 minutes and 5,795 audio excerpts of 20 endemic bird species found in the Aiguamolls de l'Emporda' Natural Park  (Gómez-Gómez et al., 2023).
This dataset was specifically created to evaluate small-footprint deep learning approaches for bird species classification using bioacoustic monitoring devices and is publicly available.

Most species had only call or song vocalizations considered, but for some species both calls and songs were obtained (Table 1). In some cases, such as the \textit{Dendrocopos minor }species, the drumming effect was selected as the identifier of the bird's presence. Similarly, for the \textit{Ciconia ciconia} species, a sound produced by a repetitive clap with their bills was labelled as bill clapping, as it was acoustically similar to woodpecker drumming but with different spectral distribution.

To automatically classify the audio events, spectrograms of the audio files were generated using a window size of one second. The manually cleaned audio files in the dataset were split into fragments of one second each, and for files shorter than one second, the vocalization was repeated to fill the window size. This window size was chosen to contain at least one complete bird vocalization and the fewest possible amount of noise. However, the duration of the audio files varied greatly among species, making it difficult to satisfy both requirements (Table 1). The mel-scale was used to emphasize the frequencies of interest.

\begin{table}[H]
\resizebox{0.5\textwidth}{!}{%
\begin{tabular}{|l|l|r|r|}
\hline
\textbf{Bird species} & \textbf{Sound type} & \textbf{Total time (sec.)} & \textbf{Number of cuts} \\ \hline
Acrocephalus arundinaceus & songs & 1982 & 453 \\ \hline
Acrocephalus melanopogon & songs & 2037 & 221 \\ \hline
Acrocephalus scirpaceus & songs & 2360 & 121 \\ \hline
Alcedo atthis & songs and calls & 351 & 418 \\ \hline
Anas strepera & songs & 292 & 96 \\ \hline
Anas platyrhynchos & songs & 229 & 70 \\ \hline
Ardea purpurea & calls & 128 & 207 \\ \hline
Botaurus stellaris & songs & 414 & 436 \\ \hline
Charadrius alexandrinus & songs and calls & 109 & 375 \\ \hline
Ciconia ciconia & bill clapping & 479 & 121 \\ \hline
Circus aeruginosus & calls & 185 & 307 \\ \hline
Coracias garrulus & calls & 178 & 267 \\ \hline
Dendrocopos minor & drumming & 563 & 494 \\ \hline
Fulica atra & calls & 123 & 372 \\ \hline
Gallinula chloropus & calls & 107 & 262 \\ \hline
Himantopus himantopus & calls & 1212 & 277 \\ \hline
Ixobrychus minutus & songs and calls & 148 & 559 \\ \hline
Motacilla flava & songs & 292 & 400 \\ \hline
Porphyrio porphyrio & songs and calls & 363 & 186 \\ \hline
Tachybaptus ruficollis & songs & 543 & 153 \\ \hline
\textbf{Total} & - & \textbf{12,096} & \textbf{5,795} \\ \hline
\end{tabular}%
}
\caption{Bird species, sound type, total time in seconds, and number of cuts.}
\label{tab:table_species}
\end{table}

\subsection{Data preprocessing}
To address the multi-class nature of this classification problem, we employed stratified sampling in dividing the dataset into training, validation, and test sets. Stratified sampling ensures that each class is properly represented in each set, which helps mitigate issues with class imbalance during model training and evaluation (Elfil \& Negida, 2017).

Our methodology sampled data points from each class in percentages of 70\%, 20\%, and 10\% for the training, validation, and test sets respectively. This stratification by class allows the model to learn from a dataset that better represents the overall distribution of classes. 

In order to properly stratify the data, we implemented a sampling strategy that accounted for the different session lengths in the dataset. The sessions varied in the number of audio samples they contained for each class. To handle this we sorted these sessions by length in descending order, so that the sessions with the most samples were first. This sorted session list allowed us to allocate the longer sessions to the training set first in order to meet the 70\% train split ratio as closely as possible. Once the training set was filled, we moved on to the validation set, and finally the test set with the remaining shorter sessions.

\begin{minipage}[c][\textheight][c]{\textwidth}
  \centering
  \makebox[\textwidth]{\includegraphics[width=1\textwidth, keepaspectratio]{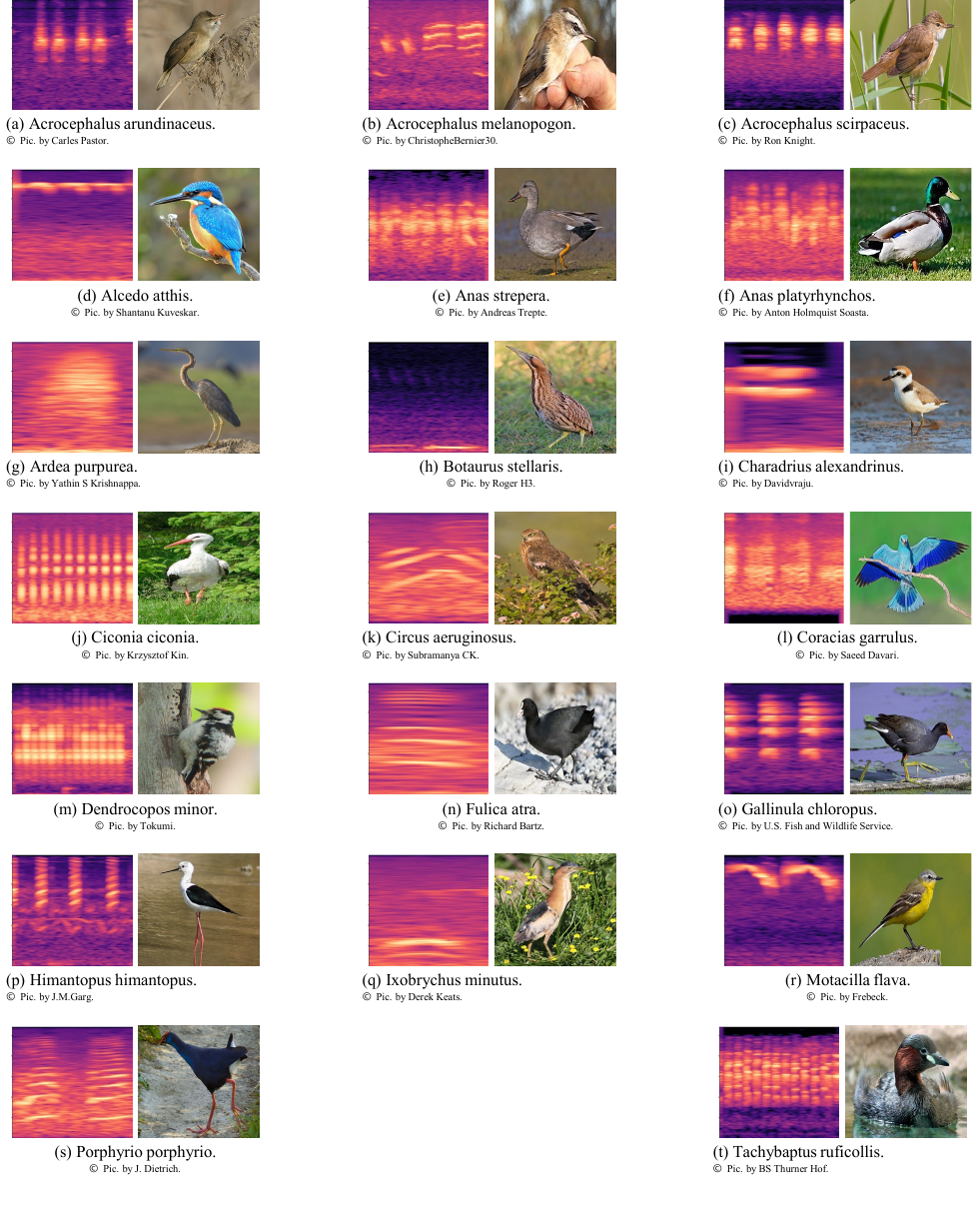}}
  \captionof{figure}{Photographs paired with spectrograms illustrating the vocalizations of various bird species, sourced from the Western Mediterranean Wetland Birds dataset as presented in the paper by (Gómez-Gómez et al., 2023)}
\end{minipage}
\twocolumn  

After sorting the sessions by length and iterating through them in order, we loaded the .npy data and labels into the corresponding train, validation, or test set. This ultimately allowed us to properly stratify the multi-class data, ensuring adequate samples of each class in each set while accounting for the variable session lengths.

To further address class imbalances within the training set, class weights were calculated to guarantee equal learning opportunities from all classes. To avoid data leakage between the sets, each audio file was strictly allocated to only one set.

To mitigate any challenges due to disparities in feature range and magnitude, data normalization was executed using the MinMaxScaler, restricting data within the range of 0 to 1. Furthermore, the training data was shuffled before being inputted into the model. This precaution was taken to prevent overfitting and enhance the model's robustness.

\begin{figure}
\centering

\begin{minipage}{0.45\textwidth}
\centering
\includegraphics[width=\linewidth]{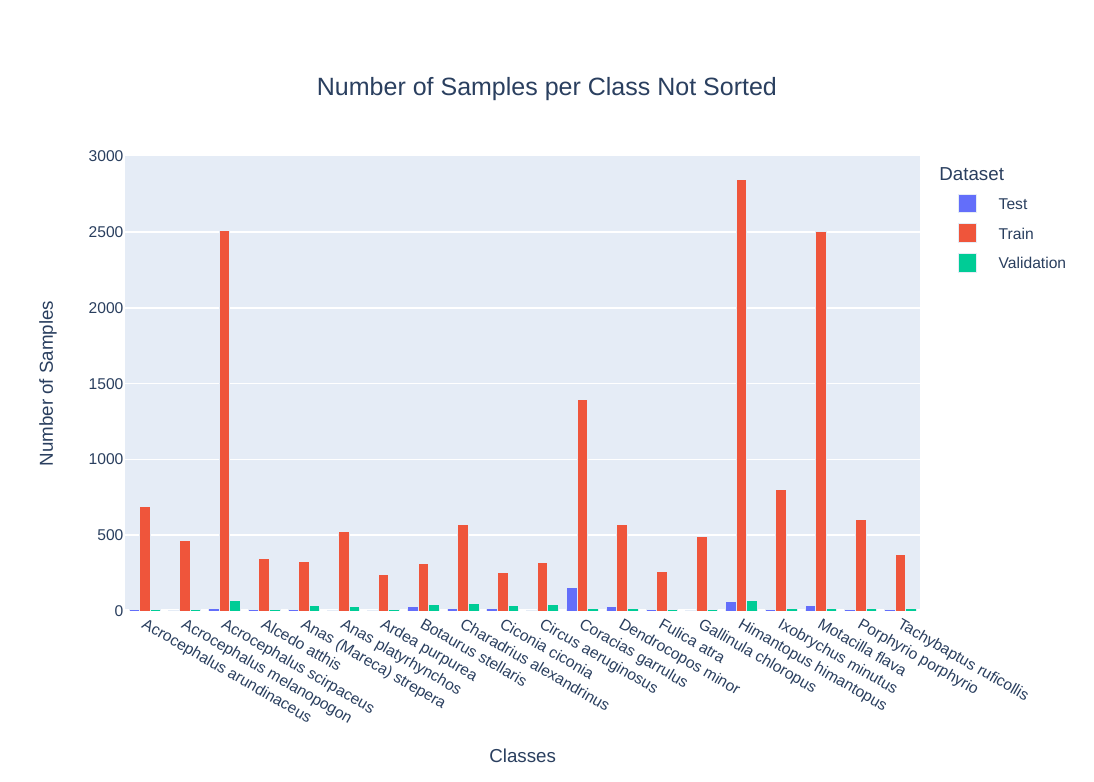}
\caption*{\small(a) unsorted }
\end{minipage}

\vspace{0.3cm} 

\begin{minipage}{0.45\textwidth}
\centering
\includegraphics[width=\linewidth]{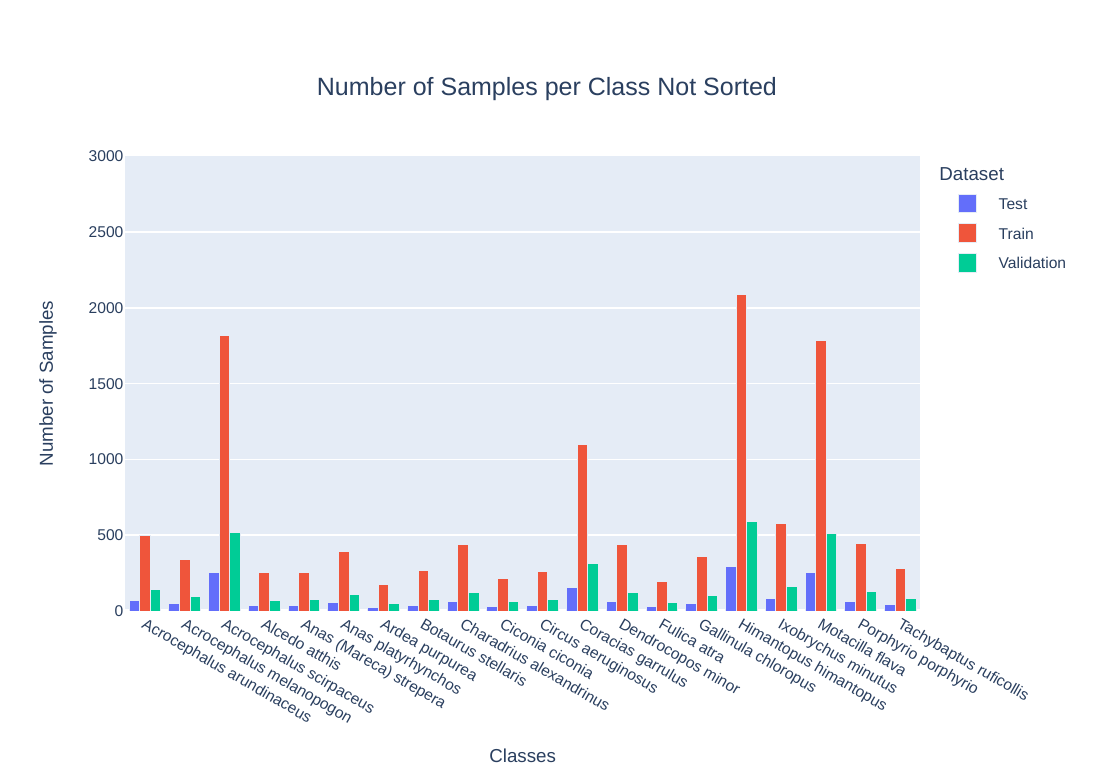}
\caption*{\small(b) sorted}
\end{minipage}

\caption{Distribution of samples across classes in the train, validation, and test sets from the Wetlands Bird Dataset. Bar heights represent sample counts, with percentages indicating the class proportion within each dataset.}
\label{fig:dataset_distribution}
\end{figure}

\subsection{Model comparison}

The first step was to acquire the three pretrained models originally used by (Gómez-Gómez et al., 2023) to analyze the dataset, namely MobileNet V2, VGG16, and ResNet50. 

\textbf{VGG16:}
Developed by the University of Oxford's Visual Geometry Group, VGG16 is a deep convolutional network known for its simplicity. It features 16 weight layers and uses small 3x3 convolution filters. Despite its power, it's computationally intensive (Simonyan and Zisserman 2014).

\textbf{ResNet50:}
ResNet50 is a deep neural network designed to tackle the vanishing gradient problem in very deep architectures. It consists of 50 layers and uses ``skip connections'' to facilitate gradient back-propagation, enabling better performance in a variety of tasks (He et al. 2016).

\textbf{MobileNet V2:}
MobileNet V2 is an efficient convolutional neural network tailored for mobile and low-resource devices. It employs inverted residual blocks and depthwise separable convolutions to reduce computational cost without sacrificing accuracy (Howard et al. 2017).

Upon securing these models, their weights were reset, and a Xavier Glorot initialization was applied to each of them. This is a standard initialization to ensure stable convergence during training (Glorot \& Bengio 2010).

Subsequently, each model was subjected to a series of 20 retraining cycles using the preprocessed data to enhance the robustness and validity of our findings. The models were trained for 500 epochs, with a batch size of 32. We used categorical cross-entropy loss, the Adam optimizer and a learning rate of 0.0001 was selected to encourage stable convergence and to avoid skipping over local minima, although small learning rate can slow down the training process.

After retraining the baseline models, our next objective entailed employing AutoKeras to explore a total of 100 distinct models, with the aim of determining the optimal architecture for our data. AutoKeras searched over various neural network architectures, including convolutional and dense networks. It automatically tuned hyperparameters like learning rate, number of layers, nodes per layer, activation functions, etc.

Upon identifying the optimal model architecture, we saved the best model. We reset its and retrained it 20 additional times using the same methodology as the baseline models. This allowed us to bolster the validity of the AutoKeras model results.

In summary, our model comparison methodology incorporated proper retraining and tuning of all models for a fair evaluation, along with AutoKeras neural architecture search to find an optimal architecture for the bird vocalization classification task.

\section{Results}
\begin{figure}[H]
\centering
\begin{minipage}{0.50\textwidth}
\centering
\includegraphics[width=1\linewidth]{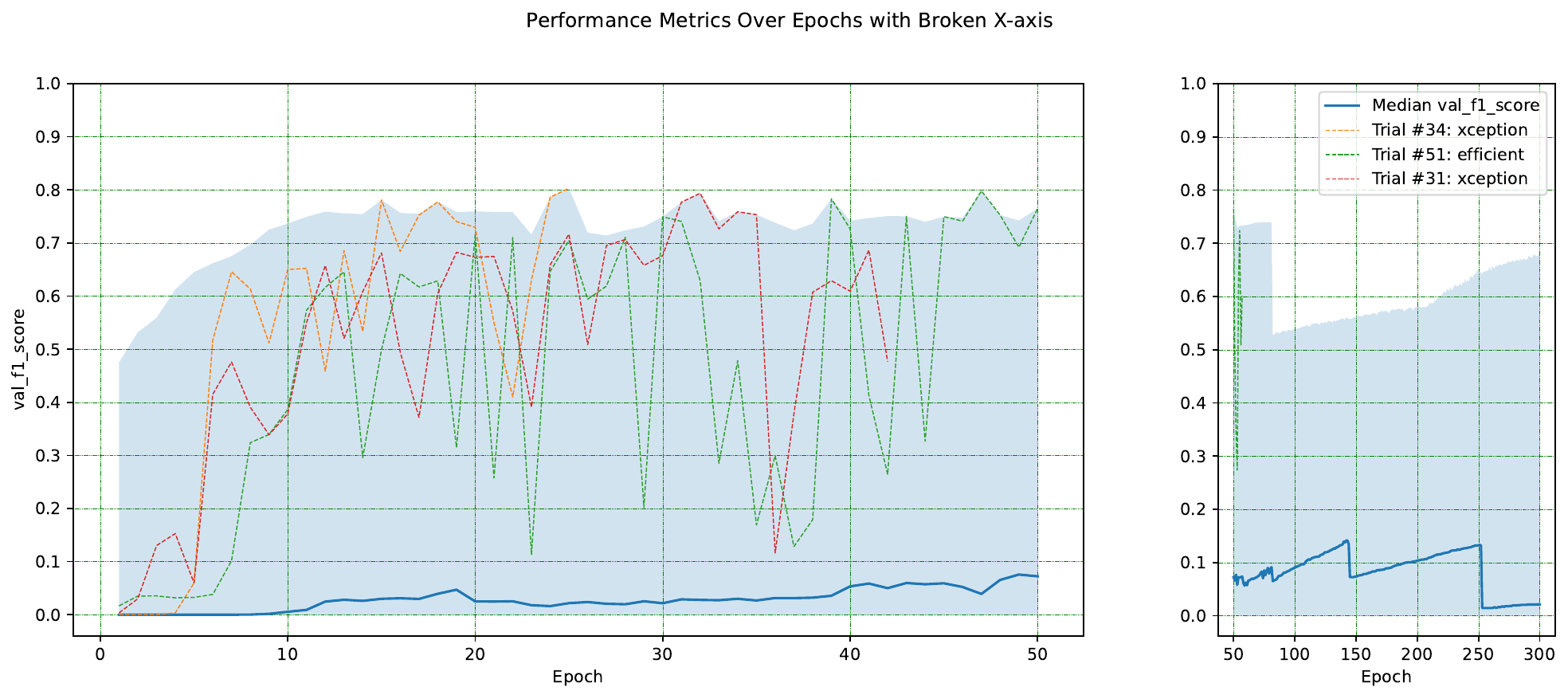}
\end{minipage}%
\caption{\label{fig:model_AutoKeras_explored} Median and top three models performances among the 100 different models trained using AutoKeras.}
\end{figure}

\begin{figure}[H]
\centering
\begin{minipage}{0.31\textwidth}
\centering
\includegraphics[width=1\linewidth]
{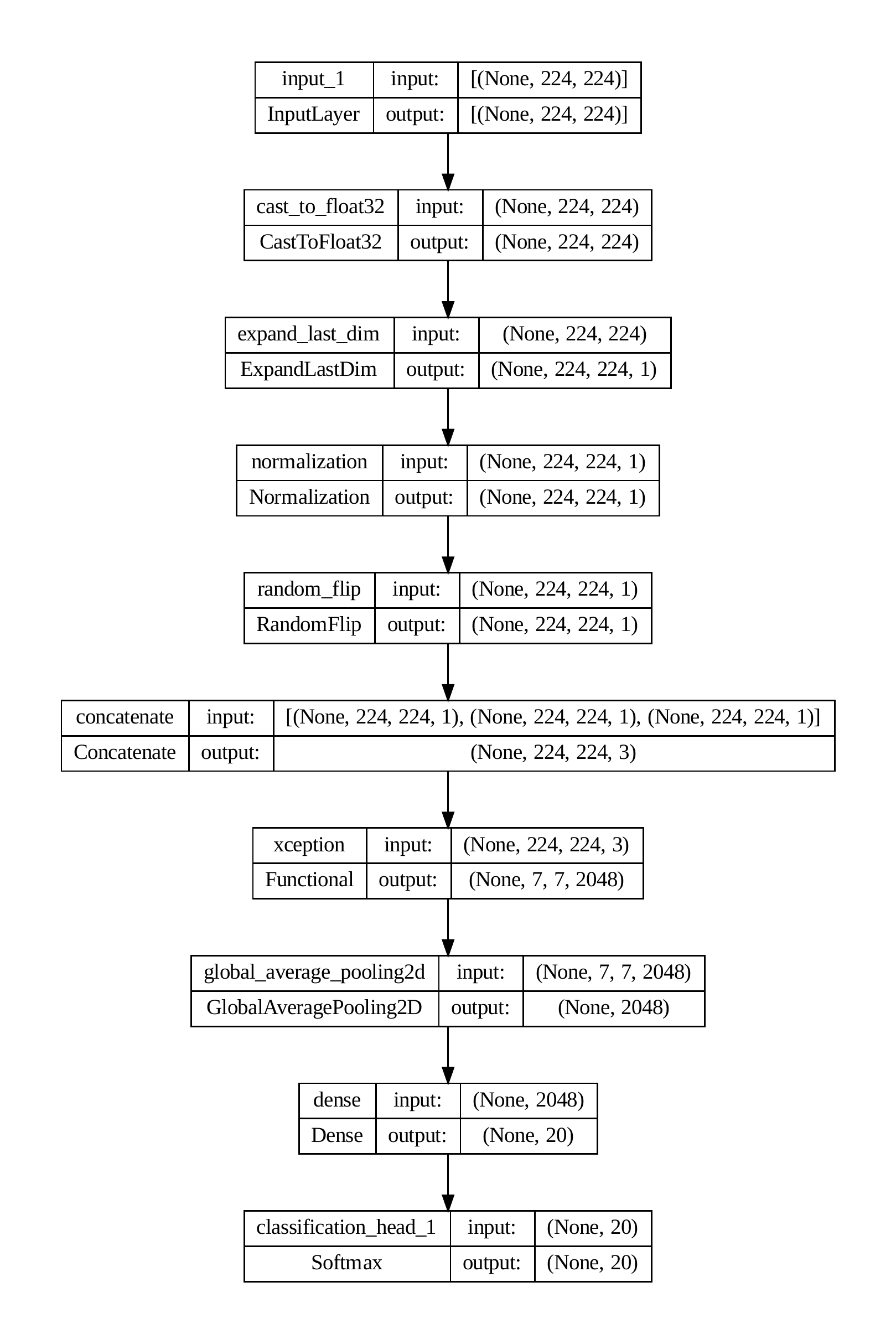}
\end{minipage}%
\caption{\label{fig:model_architecture} AutoKeras-derived Xception architecture.}
\end{figure}

The AutoKeras framework searched a total of 100 models, exploring a wide range of architectures. Figure \ref{fig:model_AutoKeras_explored} shows the variability in validation F1 scores among the models, with a median F1 score plotted over epochs. Several top performing models are highlighted, including Xception (Figure \ref{fig:model_architecture}) which demonstrated the fastest convergence.

\begin{figure}[H]
\centering
\begin{minipage}{0.50\textwidth}
\centering
\includegraphics[width=1\linewidth]{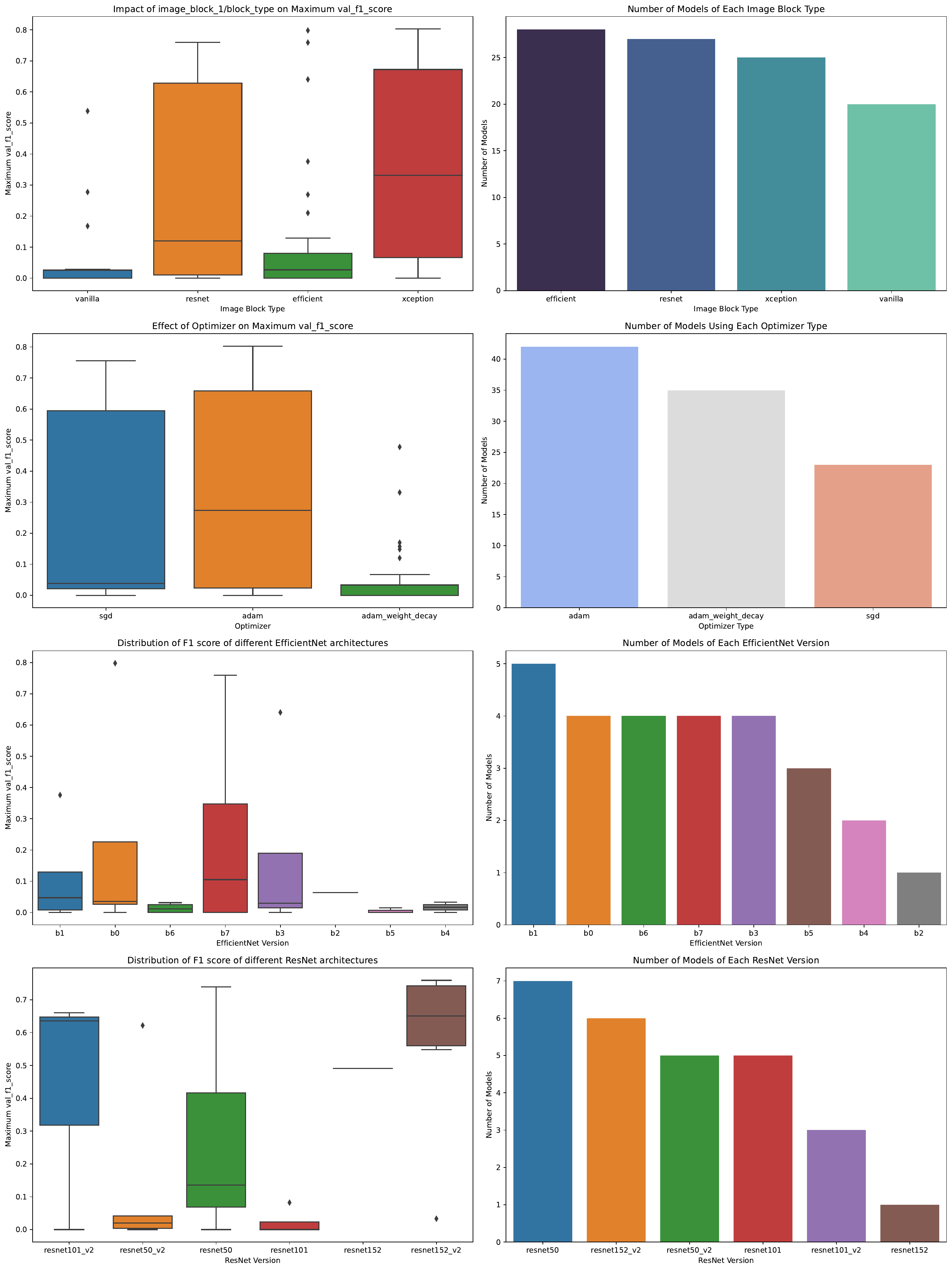}
\end{minipage}%
\caption{\label{fig:metrics} Comparative analysis of neural network architectures and optimization techniques. The left column depicts the maximum validation F1 score distributions for varying configurations, while the right column shows the number of models trained for each category.  }
\end{figure}

Figure \ref{fig:metrics} provides a more detailed analysis comparing neural network architectures and optimization techniques. The left column shows the distribution of maximum validation F1 scores for different configurations like image block type, optimizer, and specific model architectures. For example, the efficient and ResNet architectures exhibited distinct patterns in their F1 score distributions. The right column presents the number of models trained for each configuration category.

\begin{figure}
\centering
\begin{minipage}{0.20\textwidth}
\centering
\includegraphics[width=1.2\linewidth]{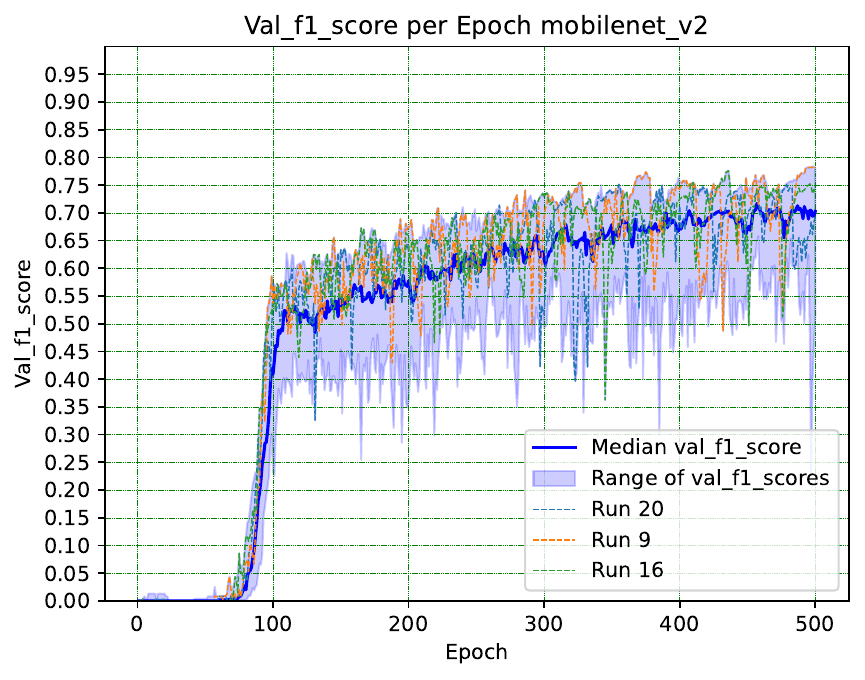}
\caption*{\small(a) mobilnet\_v2}
\end{minipage}%
\hspace{0.05\textwidth}
\begin{minipage}{0.20\textwidth}
\centering
\includegraphics[width=1.2\linewidth]{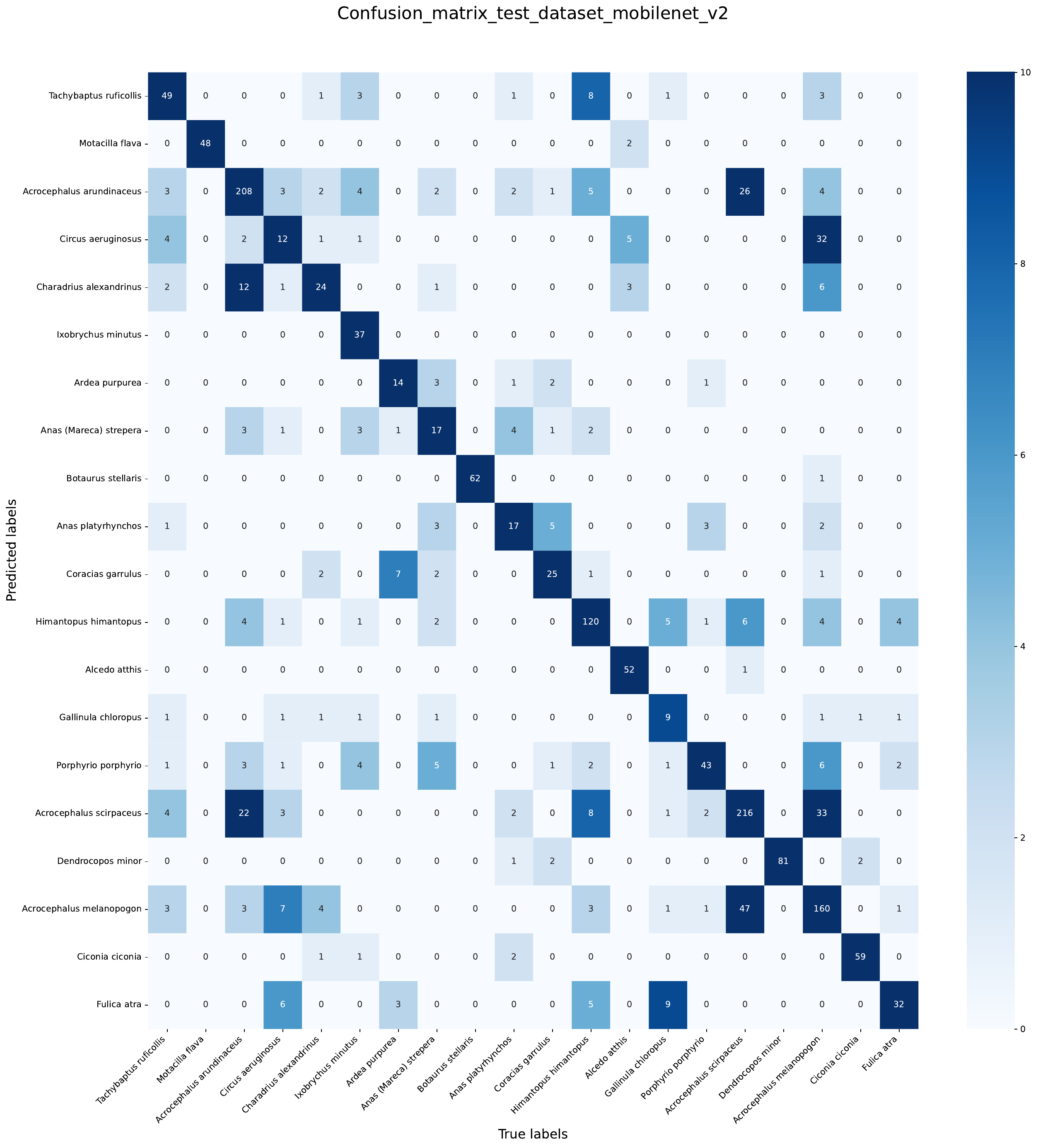}
\end{minipage}%
\hspace{0.05\textwidth}
\begin{minipage}{0.20\textwidth}
\centering
\includegraphics[width=1.2\linewidth]{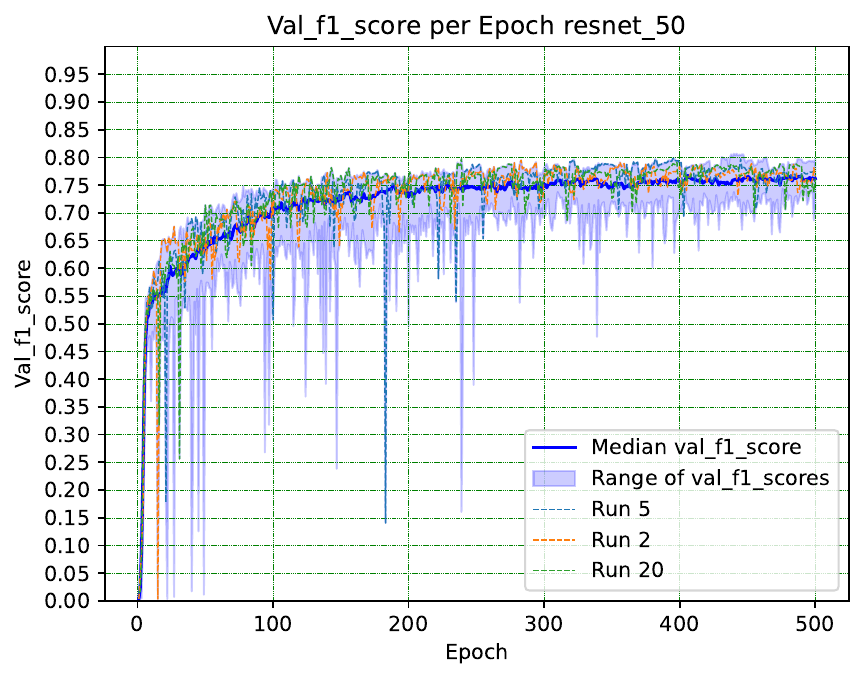}
\caption*{\small(b) resnet\_50}
\end{minipage}%
\hspace{0.05\textwidth}
\begin{minipage}{0.20\textwidth}
\centering
\includegraphics[width=1.2\linewidth]{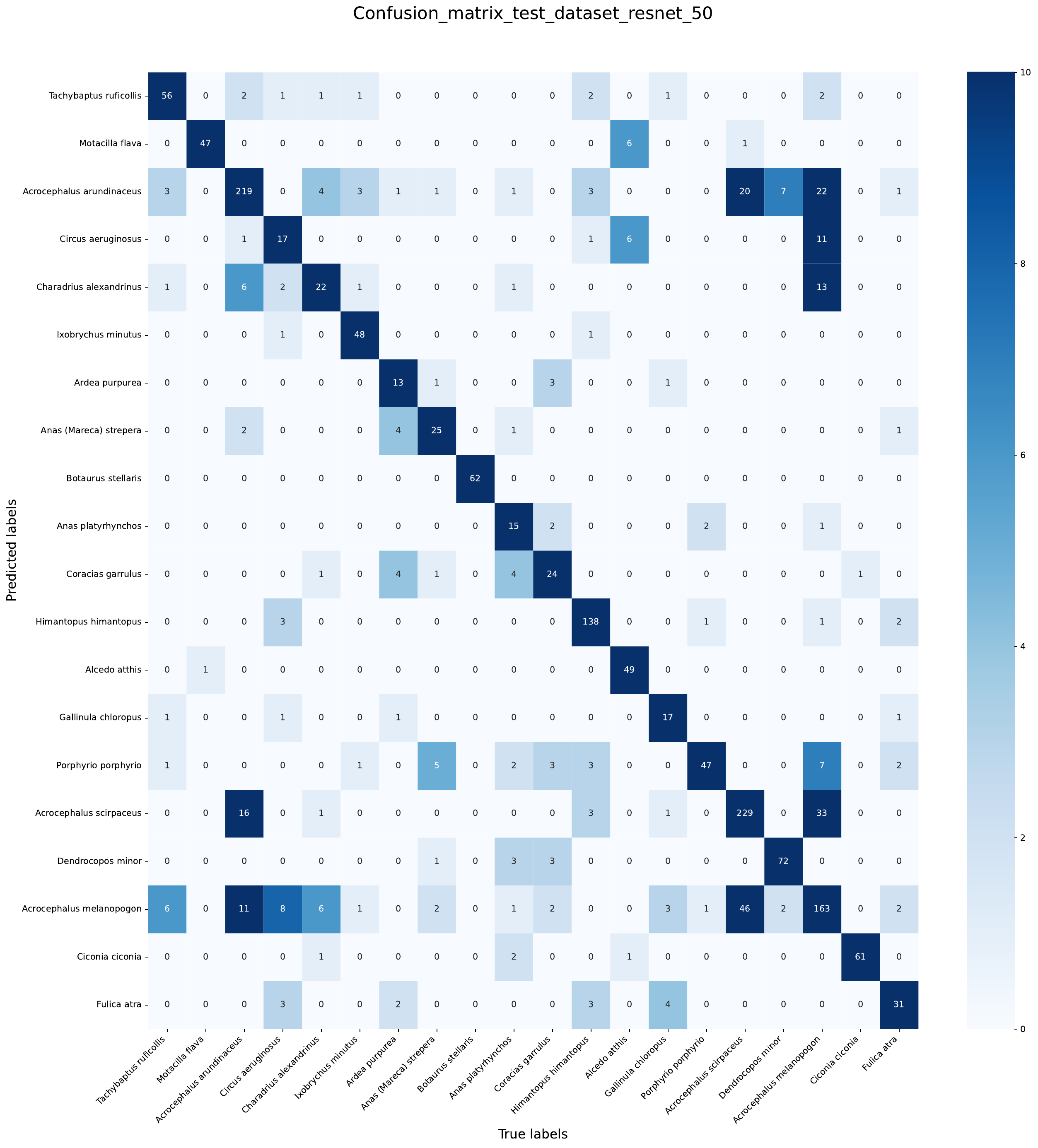}
\end{minipage}

\vspace{0.5cm}

\begin{minipage}{0.20\textwidth}
\centering
\includegraphics[width=1.2\linewidth]{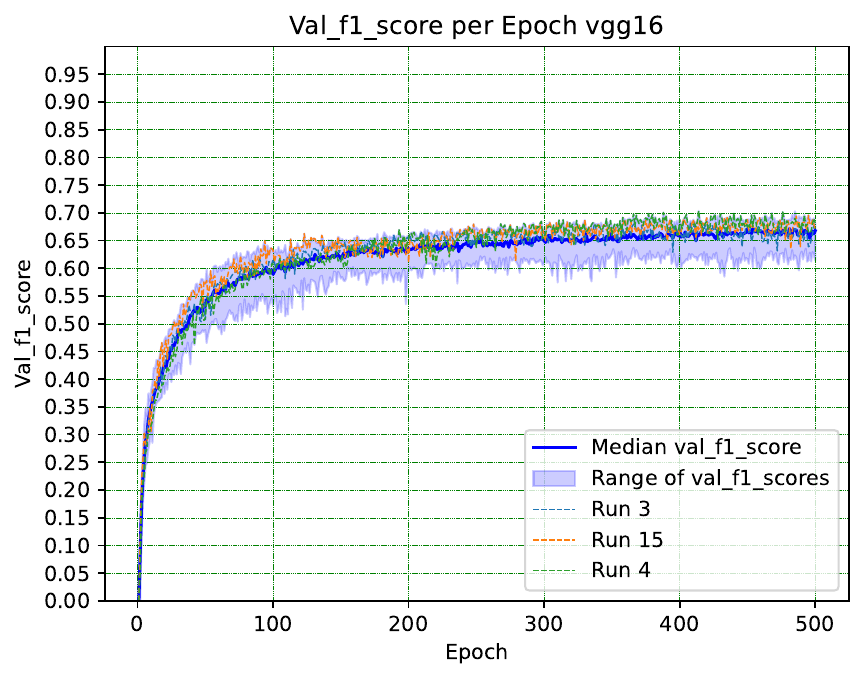}
\caption*{\small(c) vgg\_16}
\end{minipage}%
\hspace{0.05\textwidth}
\begin{minipage}{0.20\textwidth}
\centering
\includegraphics[width=1.2\linewidth]{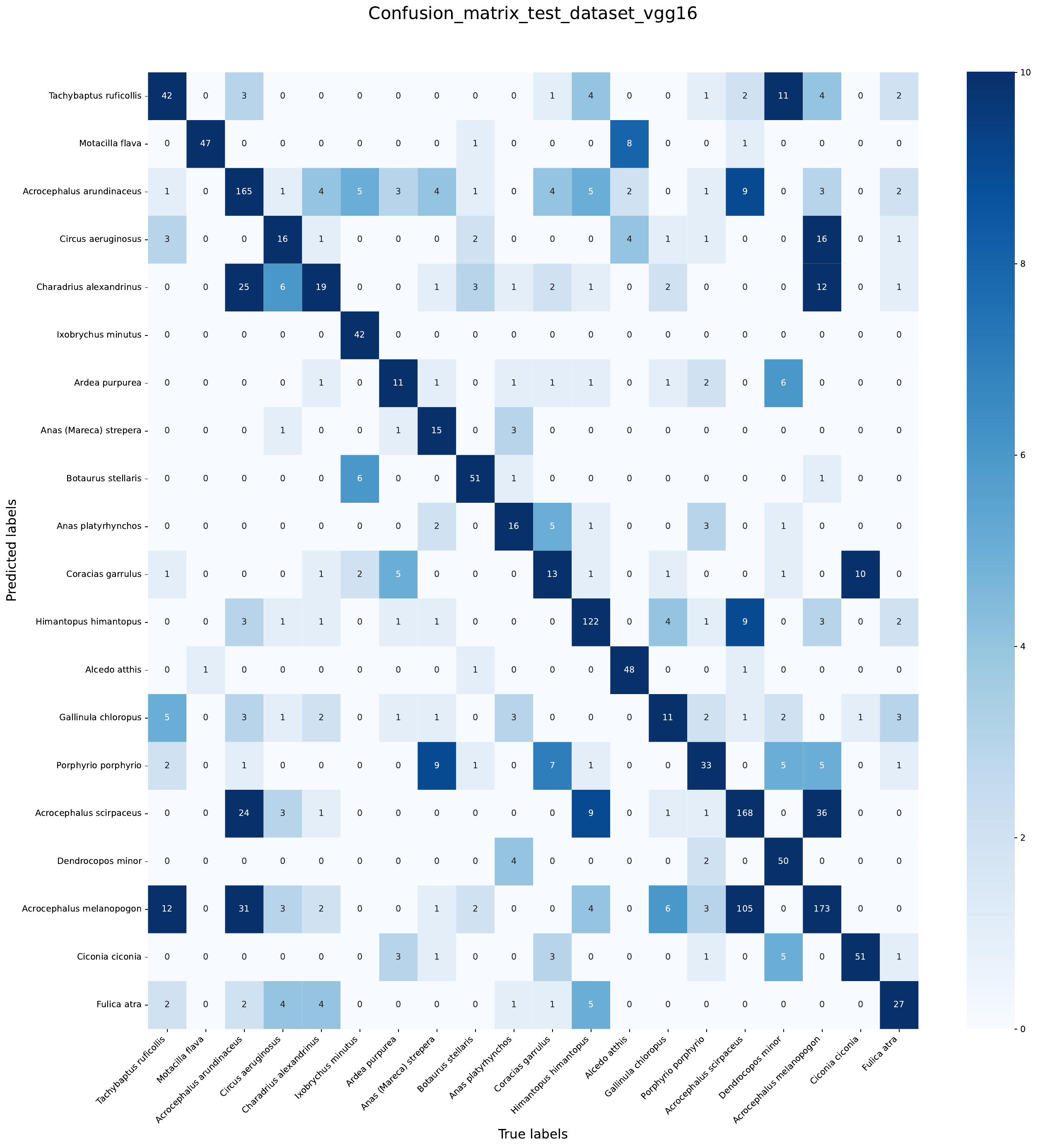}
\end{minipage}%
\hspace{0.05\textwidth}
\begin{minipage}{0.20\textwidth}
\centering
\includegraphics[width=1.2\linewidth]{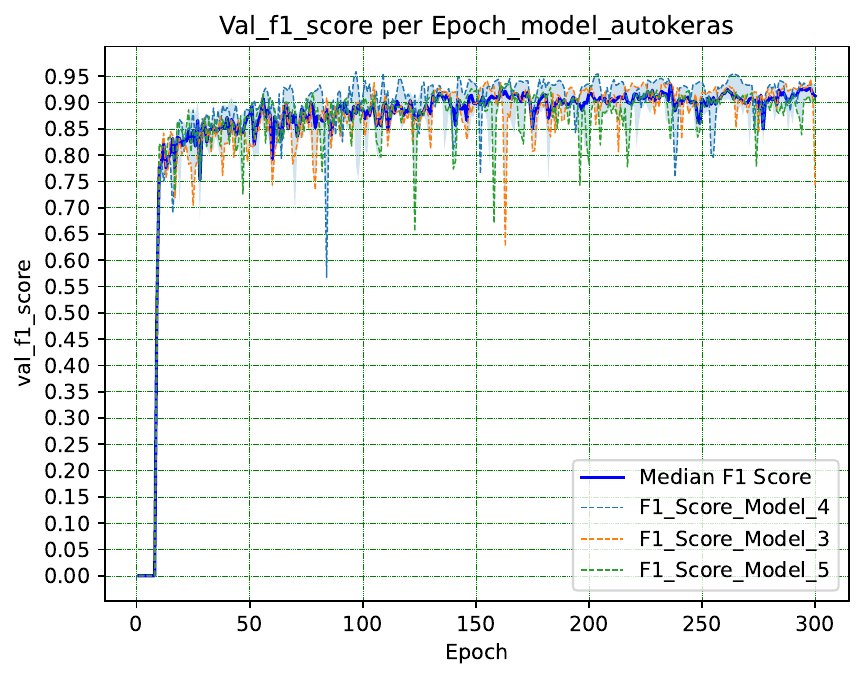}
\caption*{\small(d) Xception}
\end{minipage}%
\hspace{0.05\textwidth}
\begin{minipage}{0.20\textwidth}
\centering
\includegraphics[width=1.2\linewidth]{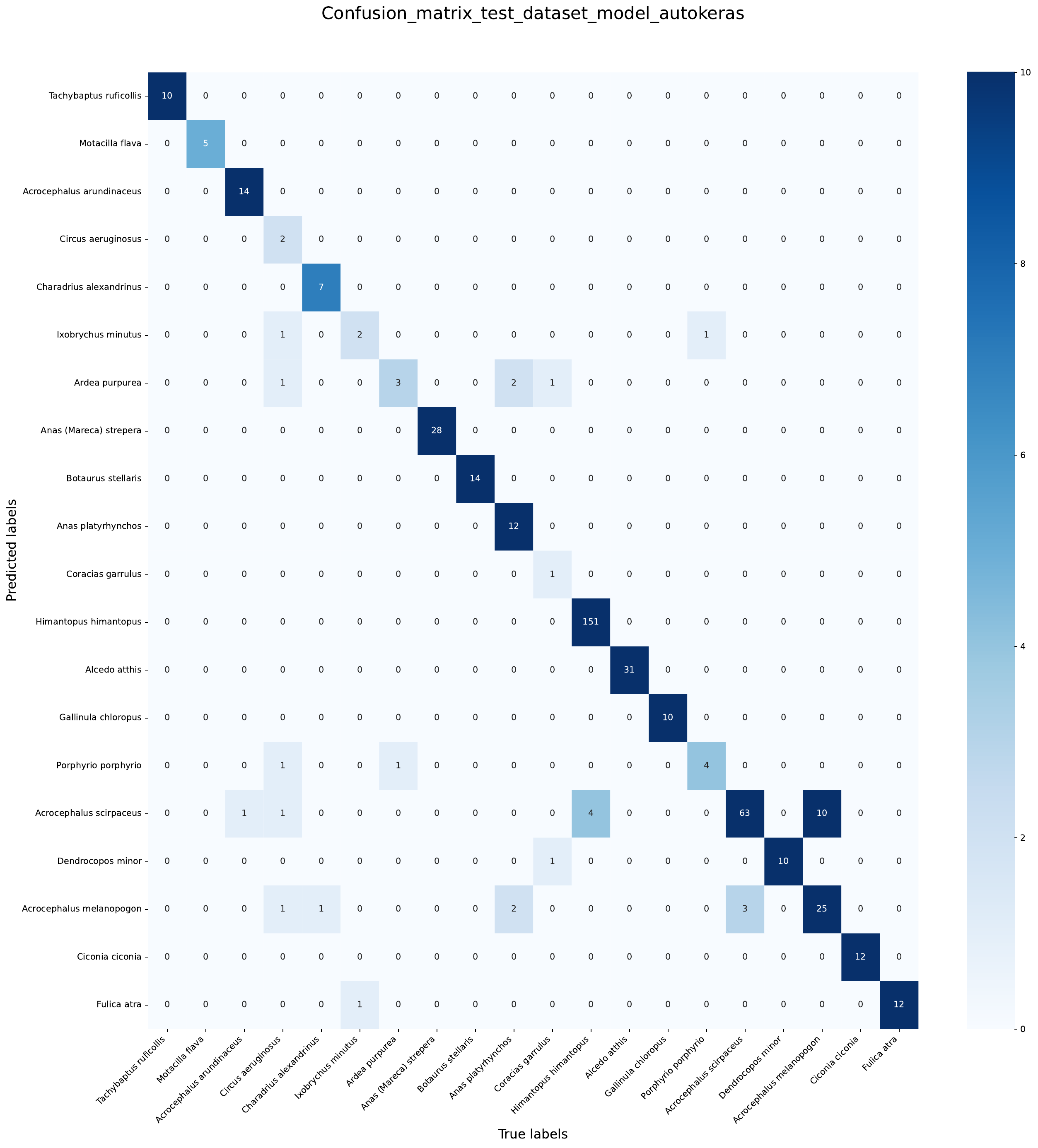}
\end{minipage}%

\vspace{1.7cm}

\resizebox{0.5\textwidth}{!}{%
\begin{tabular}{|c|c|c|c|}
\hline
\textbf{} & \textbf{Max test\_f1\_score} & \textbf{Avg test\_f1\_score} & \textbf{Min test\_f1\_score} \\
\hline
mobilenet\_v2 & 0.8230 & 0.7453 & 0.6907 \\
\hline
vgg\_16 & 0.7248 & 0.6988 & 0.6580 \\
\hline
resnet\_50 & 0.8077 & 0.7878 & 0.7641 \\
\hline
\textbf{Xception} & \textbf{0.8555} & \textbf{0.8242} & \textbf{0.7944} \\
\hline
\end{tabular}%
}
\caption{\label{fig:metrics_results} Comparison of validation F-scores (median and top three runs), confusion matrix analysis on the test dataset, and F-score evaluation on the test set. }
\end{figure}

In comparison to the three baseline models (MobileNet, ResNet50, VGG16) proposed by (Gómez-Gómez et al., 2023), the Xception model identified by AutoKeras demonstrated superior performance on both the validation and test datasets as shown in Figure \ref{fig:metrics_results}. The median and top three runs for each model are plotted over epochs, along with the confusion matrix analysis on the test set. The F1 scores on the separate test set validates that the AutoKeras model has better generalization beyond just the training and validation data. 

\begin{figure}
\vspace{0.5cm}
\centering
\begin{minipage}{0.20\textwidth}
\centering
\includegraphics[width=1.3\linewidth]{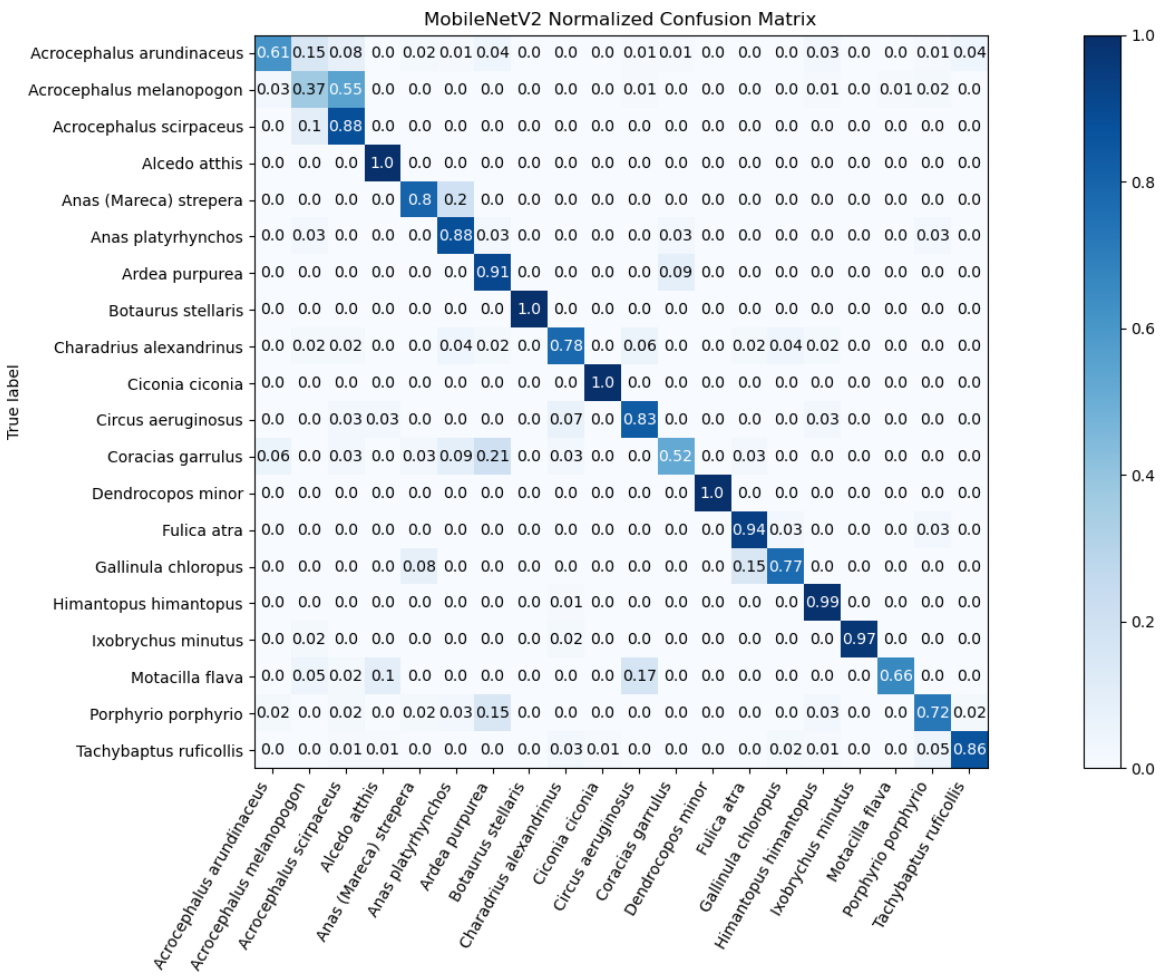}
\caption*{\small(a) mobilnet\_v2}
\end{minipage}%
\hspace{0.05\textwidth}
\vspace{-0.1cm}
\begin{minipage}{0.20\textwidth}
\centering
\includegraphics[width=1.48\linewidth]{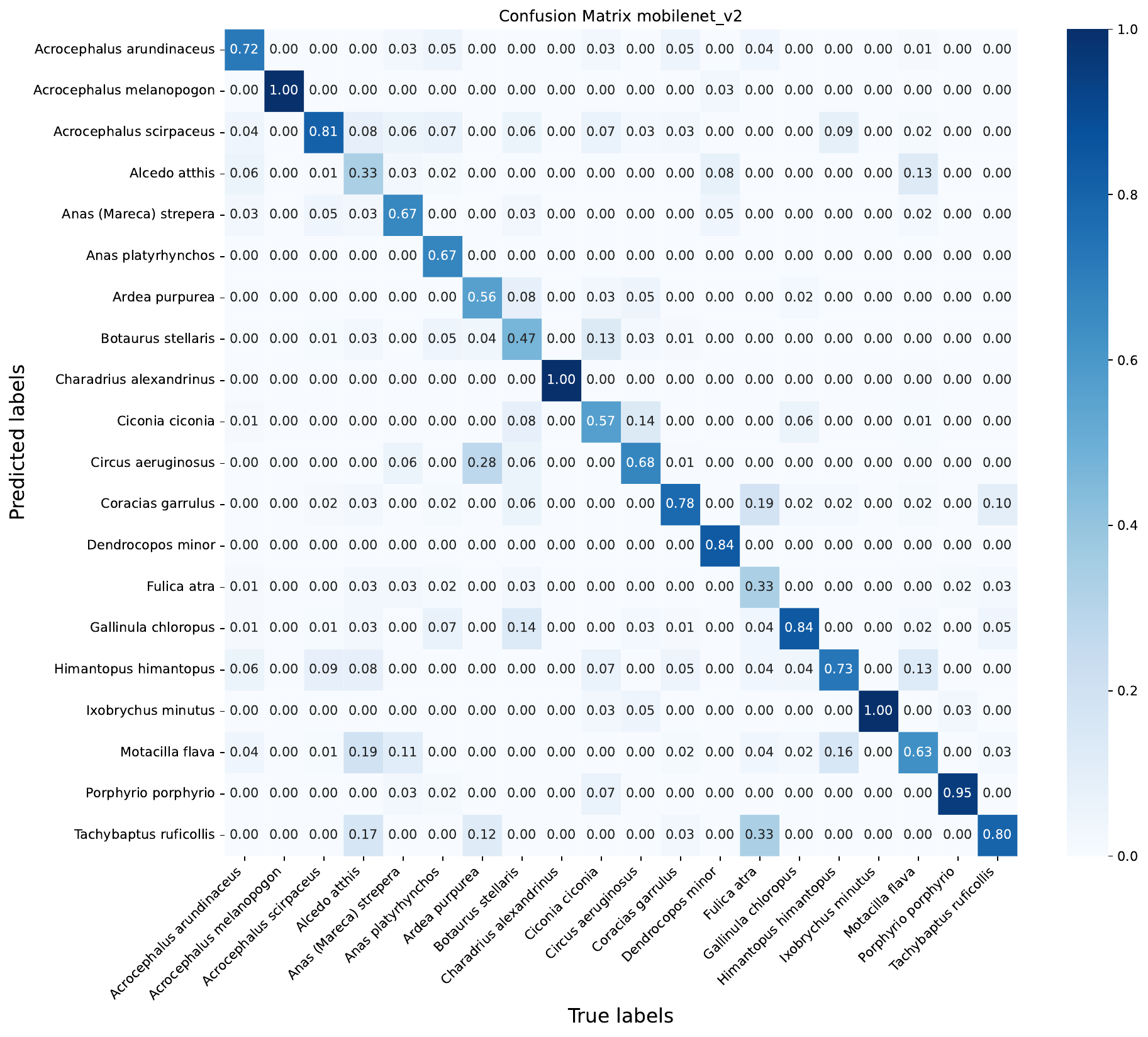}
\end{minipage}%
\hspace{0.05\textwidth}
\begin{minipage}{0.20\textwidth}
\centering
\includegraphics[width=1.3\linewidth]{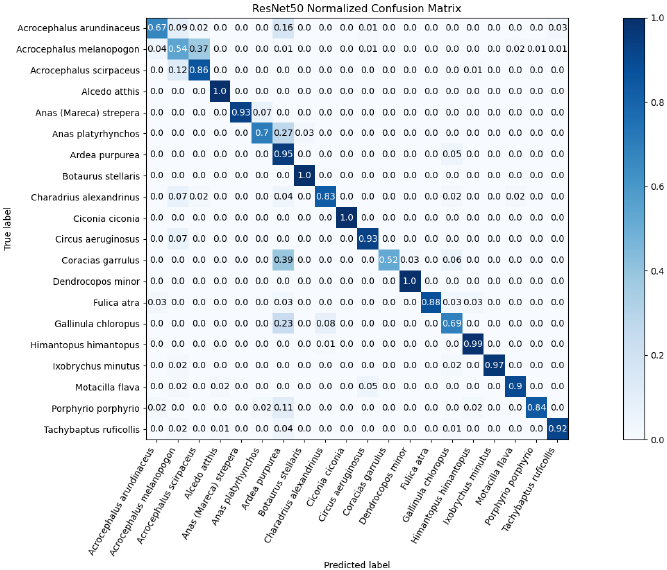}
\caption*{\small(b) resnet\_50}
\end{minipage}%
\hspace{0.05\textwidth}
\begin{minipage}{0.20\textwidth}
\centering
\includegraphics[width=1.48\linewidth]{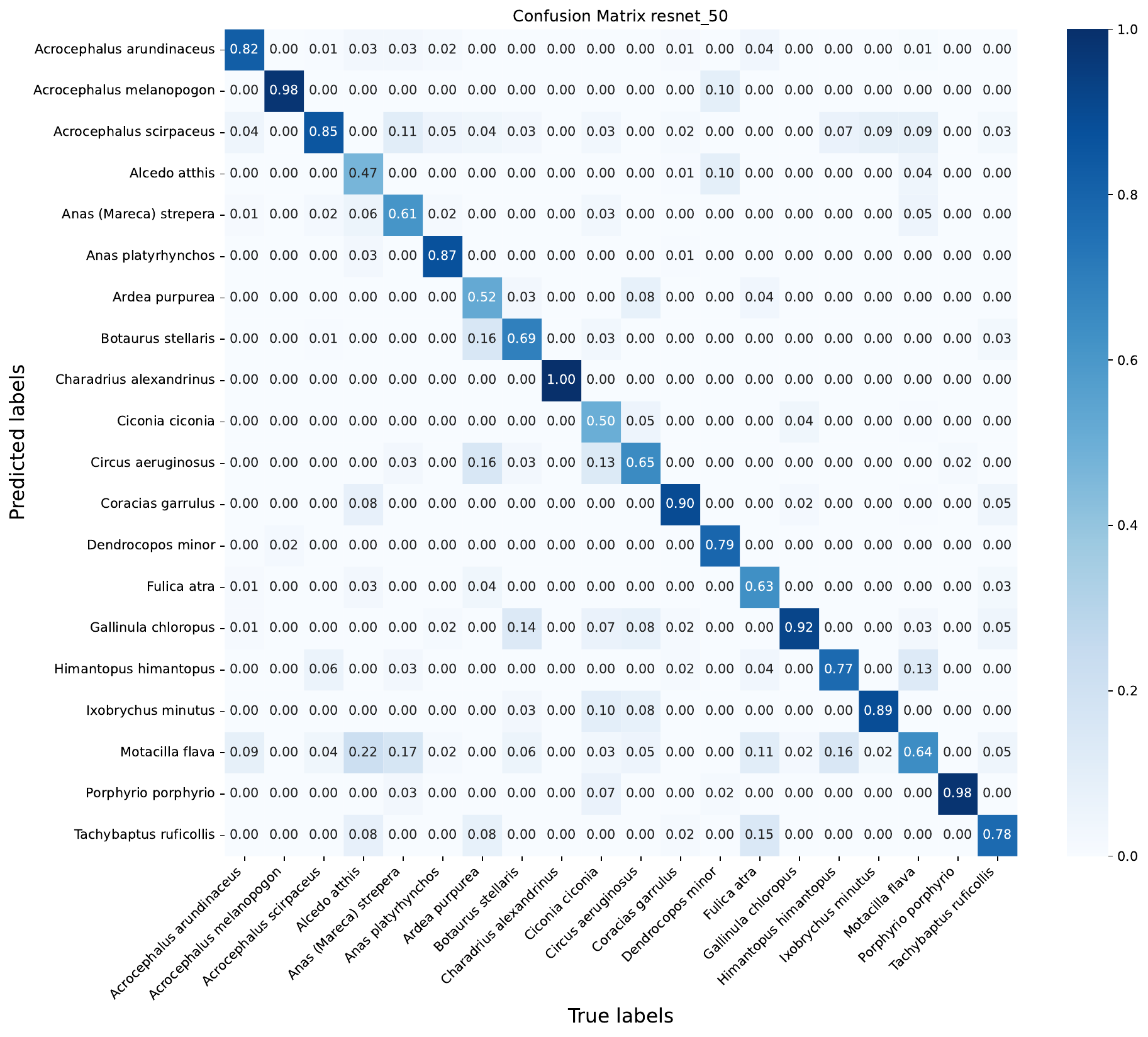}
\end{minipage}

\vspace{0.5cm}

\begin{minipage}{0.20\textwidth}
\centering
\includegraphics[width=1.3\linewidth]{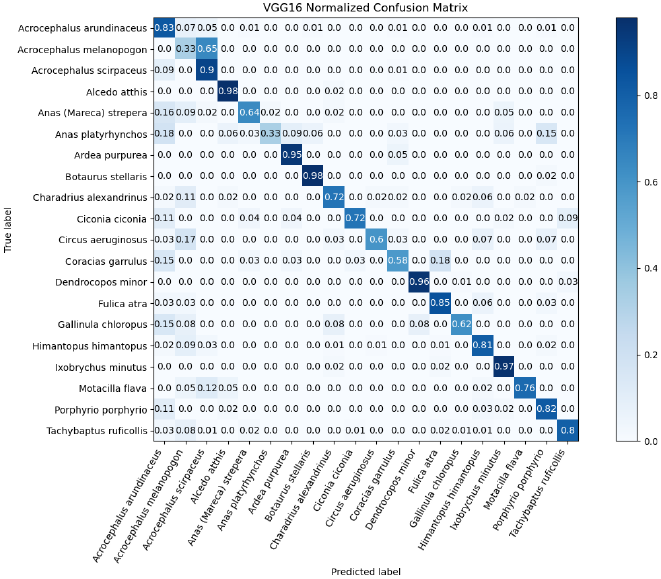}
\caption*{\small(c) vgg\_16}
\end{minipage}%
\hspace{0.05\textwidth}
\begin{minipage}{0.20\textwidth}
\centering
\includegraphics[width=1.48\linewidth]
{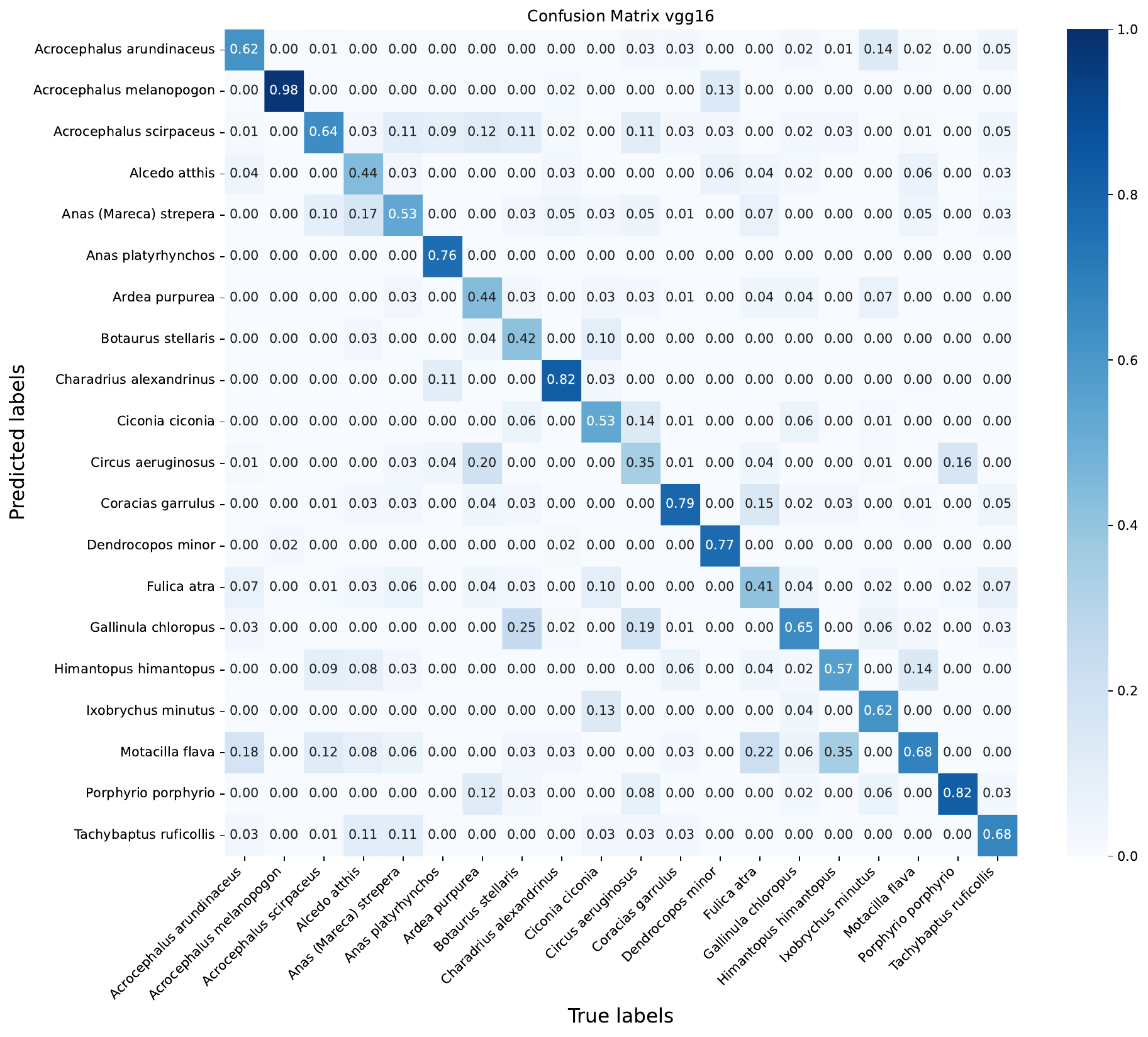}

\end{minipage}%
\hspace{0.05\textwidth}
\begin{minipage}{0.20\textwidth}
\centering
\end{minipage}%
\hspace{0.3\textwidth}
\begin{minipage}{0.20\textwidth}
\centering
\includegraphics[width=1.48\linewidth]{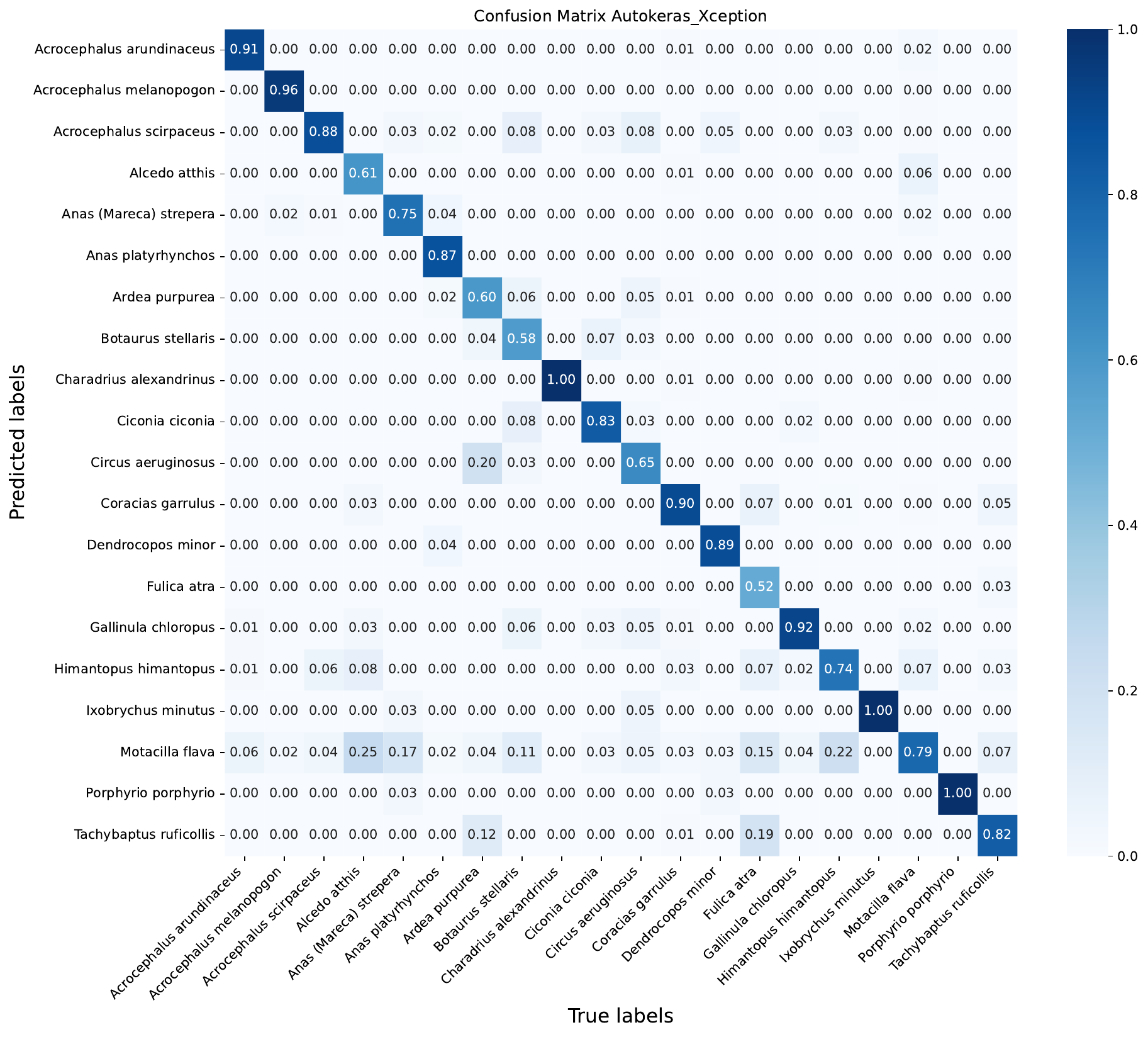}
\end{minipage}%

\vspace{0.98cm}

\resizebox{0.55\textwidth}{!}{%
\begin{tabular}{|c|c|c|c|c|}
\hline
\textbf{} & \textbf{F1-score (Gómez-Gómez et al., 2023)} & \textbf{Max test\_f1\_score} & \textbf{Avg test\_f1\_score} & \textbf{Min test\_f1\_score} \\
\hline
    mobilenet\_v2 & 0.789 & 0.8230 & 0.7453 & 0.6907 \\
    \hline
    vgg\_16 & 0.768 & 0.7248 & 0.6988 & 0.6580 \\
    \hline
    resnet\_50 & 0.834 & 0.8077 & 0.7878 & 0.7641 \\
\hline
\textbf{Xception} & \textbf{-} & \textbf{0.8555} & \textbf{0.8242} & \textbf{0.7944} \\
\hline
\end{tabular}%
}
\caption{\label{fig:confusion_matrices} Comparison between our result (right) and result obtained by (Gómez-Gómez et al., 2023) (left) of validation F-scores (median and top three runs), confusion matrix analysis on the test dataset, and F-score evaluation on the test set.}
\end{figure}

Confusion matrix analysis was conducted for all four models, revealing certain shared challenges among them (Figure \ref{fig:confusion_matrices}). Across all models, it was observed that some classes were consistently misclassified, suggesting that these classes may share similar features that confuse the models, or alternatively, the data for these classes may be limited or of lower quality. This is a common issue in multi-class classification tasks and indicates an area for future investigation.

In summary, our results demonstrated the superiority of the AutoKeras-derived Xception model over the pretrained models in terms of performance on both validation and test datasets.

\begin{figure}[H]
\centering
\begin{minipage}{0.50\textwidth}
\centering
\includegraphics[width=1\linewidth]{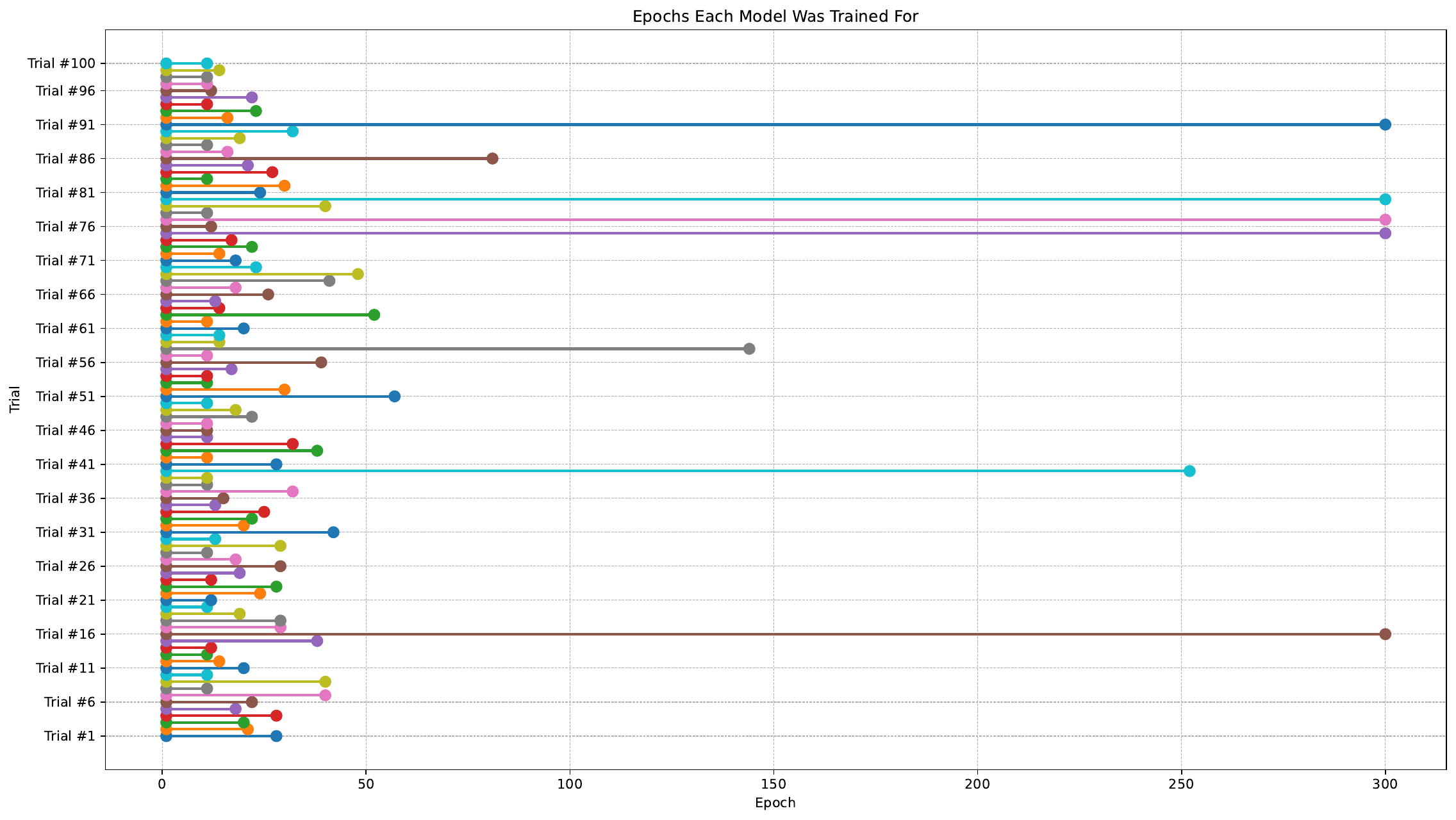}
\end{minipage}%
\caption{\label{fig:lifespan} Number of epochs for which each model was trained for by AutoKeras.}
\end{figure}

\begin{figure}[H]
\centering
\begin{minipage}{0.50\textwidth}
\centering
\includegraphics[width=1\linewidth]{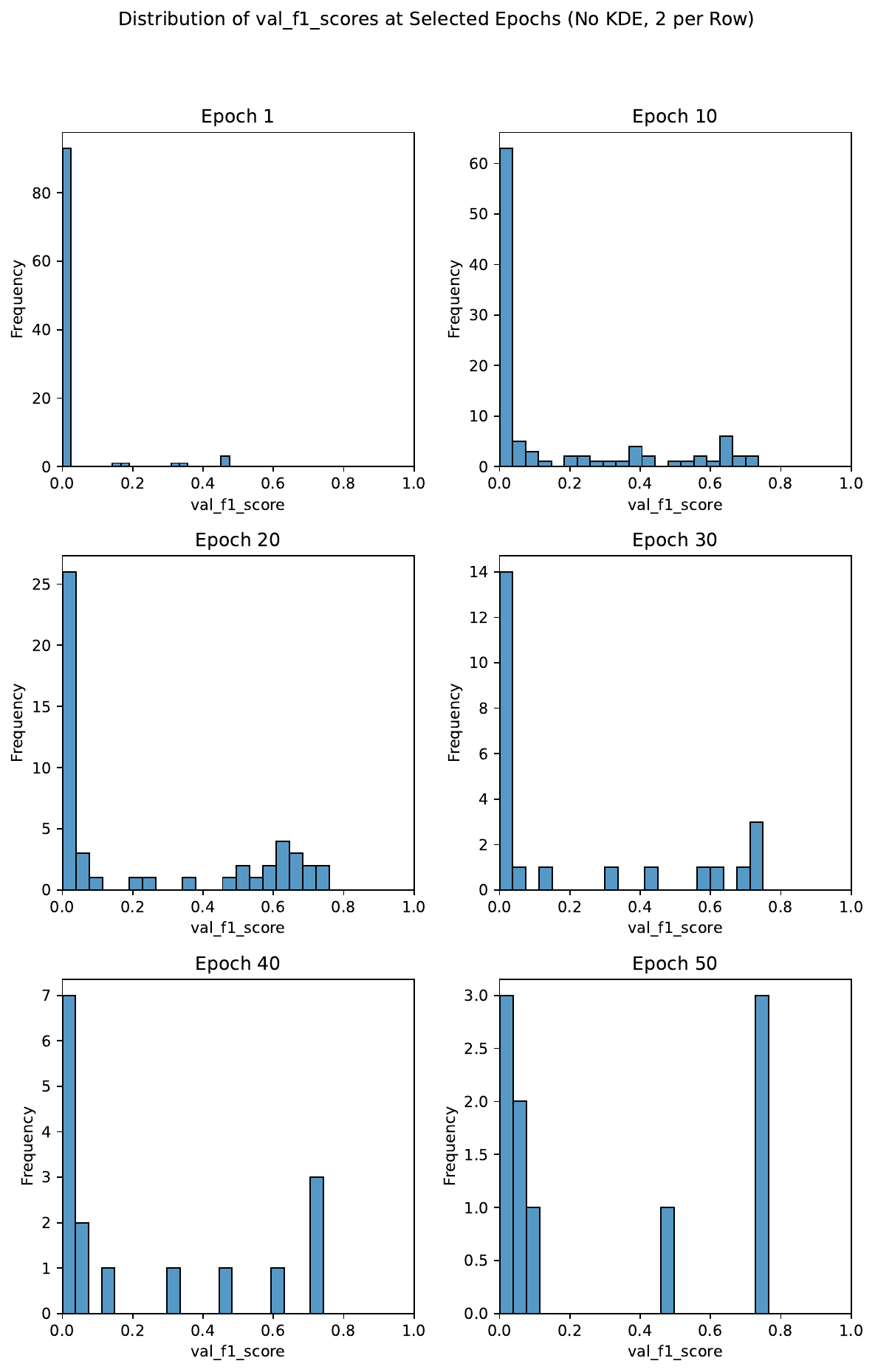}
\end{minipage}%
\caption{\label{fig:histrograms_f_score} Distribution of Validation F-Scores of each model across selected epochs.}
\end{figure}

\newpage

\section{Discussion}
The original authors, Gómez-Gómez et al. (2023), used a stratified sampling technique based on session length. However, they overlooked variations in session durations within the dataset. Figure \ref{fig:dataset_distribution} illustrates this oversight: even though sessions are split 70\%, 20\%, and 10\%, the differing session lengths result in an imbalanced representation. Our data split, which accounts for session lengths, significantly deviates from their method. This difference is evident in our results, particularly in the confusion matrix analysis and overall model performance. It underscores the crucial role of data preprocessing for unbalanced datasets. Without proper stratification, models risk poor generalization and inaccurate evaluations.

The methodology of automated machine learning involves the comparative analysis of several models, subsequently determining the optimal one based on its performance on the validation set. Hence, the necessity for a separate test set becomes clear, reinforcing the importance of model generalization beyond the confines of the training and validation data.

It is crucial to highlight the importance of accurately representing both the number of samples and misclassified instances in the confusion matrix. This approach provides a clearer understanding of how the dataset was split, aligning with the principles of the open science community.  This gives a more detailed picture of system performance than in (Gómez-Gómez et al., 2023), for example, who reported percentages.

\section{Conclusion}
Our research demonstrates that the model derived via AutoKeras outperformed the three models initially proposed by the author, as shown in Figure \ref{fig:metrics_results}. This indicates the potential of automated machine learning as a transformative tool that reduces the need for manual design and optimisation of neural network models, representing an exciting trajectory for future research in this area.

Furthermore, our results highlight the impact of using stratified sampling  strategy that accounted for the different session lengths in the dataset when working with multi-class imbalanced datasets. By ensuring proper representation of all classes in the train, validation, and test sets, we were able to achieve superior performance compared to the original authors. This underscores the need to report sampling methodology in research studies, as suboptimal sampling can lead to poor model generalization and inaccurate evaluation.

Additionally, we emphasize the value of presenting full confusion matrix numbers rather than solely percentages when dealing with class imbalance. Reporting the actual misclassification counts provides greater insight into model performance and aids reproducibility. The inconsistencies between our stratified approach and the original authors' non-stratified methodology demonstrates the significant impact sampling can have on results.

Overall, this study illuminates automated machine learning and stratified sampling as critical tools for advancing bioacoustics research and models. Our methodology and reporting practices align with open science principles and could help improve model development, evaluation, and reproducibility in this growing field. The code and models are openly available to facilitate further research.

\section*{Acknowledgements}
Funding and publishing assistance provided by Tilburg University.

Giulio Tosato: Project ideation, dataset preprocessing, AutoKeras implementation, trained  VGG16, ResNet50 and Mobilnet\'V2, data visualizations, rewrote parts of paper.

Pramatya Jati: reviewed and adjusted paper, handled dataset, rewrote parts of paper.

Abdelrahman Shehata: reviewed and adjusted paper, rewrote parts of paper, augmented audio files.

Kees Kamp: reviewed and adjusted paper, handled dataset.

Joshua Janssen: reviewed and adjusted paper, augmented audio files.

Dan Stowell: advised on study design, reviewed and edited the paper.

\section*{Declaration of Interest}
 The authors declare that they have no known competing financial interests or personal relationships that could have appeared to influence the work reported in this paper.
 
\section*{Data availability}
Dataset available from \\
\url{https://zenodo.org/records/7505820} \\
Deep learning models available at \\
\url{https://github.com/giuliotosato/AutoKeras-bioacustic}

\section*{Bibliography}
Brown, A., Montgomery, J., \& Garg, S. (2021). Automatic construction of accurate bioacoustics workflows under time constraints using a surrogate model. Applied Soft Computing, 113, 107944. \url{https://doi.org/10.1016/j.asoc.2021.107944}

Elfil, M., \& Negida, A. (2017). Sampling methods in Clinical Research; an Educational Review. PubMed, 5(1), e52. \url{https://pubmed.ncbi.nlm.nih.gov/28286859}

Glorot, X. ,Bengio, Y.. (2010). Understanding the difficulty of training deep feedforward neural networks. Proceedings of the Thirteenth International Conference on Artificial Intelligence and Statistics, in Proceedings of Machine Learning Research 9:249-256 Available from   
\url{https://proceedings.mlr.press/v9/glorot10a.html}.

Gomez, J. (2022). BirdMLClassification [Python]. \url{https://github.com/jogomez97/BirdMLClassification}

Gómez-Gómez, J., Vidaña-Vila, E., \& Sevillano, X. (2023). Western Mediterranean Wetland Birds dataset: A new annotated dataset for acoustic bird species classification. Ecological Informatics, 75, 102014.  \url{https://doi.org/10.1016/j.ecoinf.2023.102014 }

K. He, X. Zhang, S. Ren and J. Sun, "Deep Residual Learning for Image Recognition," \textit{2016 IEEE Conference on Computer Vision and Pattern Recognition (CVPR)}, Las Vegas, NV, USA, 2016, pp. 770-778, doi: 10.1109/CVPR.2016.90.  \url{
https://ieeexplore.ieee.org/document/7780459
}.

Howard AG, Zhu M, Chen B, Kalenichenko D, Wang W, Weyand T, Andreetto M, Adam H. (2017). MobileNets: Efficient Convolutional Neural Networks for Mobile Vision Applications. arXiv:1704.04861
\url{https://arxiv.org/abs/1704.04861}

Jin, H., Song, Q., \& Hu, X. (2019). Auto-Keras: An Efficient Neural Architecture Search System. Knowledge Discovery and Data Mining.  \url{https://doi.org/10.48550/arXiv.1806.10282}.

Simonyan K, Zisserman A. 2014. Very deep convolutional networks for large-scale image recognition. arXiv preprint arXiv:1409.1556.\url{https://doi.org/10.48550/arXiv.1409.1556}

Stowell, D. (2022). Computational bioacoustics with deep learning: a review and roadmap. PeerJ, 10, e13152. \url{https://doi.org/10.7717/peerj.13152}

Truong, A. D., Walters, A., Goodsitt, J., Hines, K. E., Bruss, C. B., \& Farivar, R. (2019). Towards Automated Machine Learning: Evaluation and Comparison of AutoML Approaches and Tools. ArXiv (Cornell University). \url{https://doi.org/10.1109/ictai.2019.00209}

Tuggener, L., Amirian, M., Rombach, K., L\"{o}rwald, S., Varlet, A., Westermann, C., \& Stadelmann, T. (2019). Automated Machine Learning in Practice: State of the Art and Recent Results. 2019 6th Swiss Conference on Data Science (SDS), 31–36. \url{https://doi.org/10.1109/SDS.2019.00-11}

Wei, T. (2016). Network morphism. \url{https://doi.org/10.48550/arXiv.1603.01670}

\end{document}